\documentclass{article}

\usepackage[preprint]{neurips_2025}

\usepackage[utf8]{inputenc}
\usepackage[T1]{fontenc}
\usepackage{hyperref}
\usepackage{url}
\usepackage{booktabs}
\usepackage{amsfonts}
\usepackage{amsmath}
\usepackage{amssymb}
\usepackage{nicefrac}
\usepackage{microtype}
\usepackage{graphicx}
\usepackage{xcolor}
\usepackage{multirow}
\usepackage{caption}
\usepackage{subcaption}
\usepackage{array}
\usepackage{makecell}

\definecolor{tierone}{HTML}{2E7D32}
\definecolor{tiertwo}{HTML}{F57F17}
\definecolor{tierfail}{HTML}{C62828}

\title{Behavioral Steering in a 35B MoE Language Model via SAE-Decoded Probe Vectors: \\One Agency Axis, Not Five Traits}

\author{
  Jia Qing Yap \\
  Independent Researcher \\
}

\begin{document}

\maketitle

\begin{abstract}
We train nine sparse autoencoders (SAEs) on the residual stream of Qwen~3.5-35B-A3B, a 35-billion-parameter Mixture-of-Experts model with a hybrid GatedDeltaNet/attention architecture, and use them to identify and steer five agentic behavioral traits.
Our method trains linear probes on SAE latent activations, then projects the probe weights back through the SAE decoder to obtain continuous steering vectors in the model's native activation space: $\mathbf{v} = W_{\text{dec}}^\top \mathbf{w}_{\text{probe}}$.
This bypasses the SAE's top-$k$ discretization, enabling fine-grained behavioral intervention at inference time with no retraining.

Across 1,800 agent rollouts (50 scenarios $\times$ 36 conditions), we find that autonomy steering at multiplier $\alpha = 2$ achieves Cohen's $d = 1.01$ ($p < 0.0001$; peak $d = 1.04$ at $\alpha = 3$), shifting the model from asking the user for help 78\% of the time to proactively executing code and searching the web.
Cross-trait analysis, however, reveals that all five steering vectors primarily modulate a single dominant \emph{agency axis} (the disposition to act independently versus defer to the user), with trait-specific effects appearing only as secondary modulations in tool-type composition and dose-response shape.

The most informative result is a dissociation between two traits probed from the same SAE.
Risk calibration and tool-use eagerness share the same SAE and layer with nearly identical probe $R^2$ (0.795 vs.\ 0.792), yet their probe vectors are nearly orthogonal (cosine similarity $= -0.017$).
The tool-use vector steers behavior ($d = 0.39$); the risk-calibration vector produces only suppression.
Probe $R^2$ measures correlation, not causal efficacy.
A predictive direction in latent space need not be causally upstream of behavior.
We additionally show that steering only during autoregressive decoding has zero effect ($p > 0.35$), providing causal evidence that behavioral commitments are computed during prefill in GatedDeltaNet architectures.

Code, trained SAEs, and steering vectors are publicly available.\footnote{\url{https://github.com/zactheaipm/qwenscope}} \footnote{\url{https://huggingface.co/zactheaipm/qwen35-a3b-saes}}
\end{abstract}

%=============================================================================
\section{Introduction}
\label{sec:intro}
%=============================================================================

A central goal of mechanistic interpretability is to move from \emph{observing} what language models do to \emph{intervening} on how they do it.
Sparse autoencoders (SAEs) have emerged as a promising tool for decomposing model representations into interpretable features \citep{cunningham2023sparse, bricken2023monosemanticity}, and recent work has demonstrated that clamping or amplifying individual SAE features can steer model behavior \citep{templeton2024scaling}.
Separately, activation engineering methods add learned directions to the residual stream to shift model outputs without retraining \citep{turner2023activation, li2023inference}.

These approaches have primarily been demonstrated on models up to 7B parameters with standard transformer architectures.
Whether they extend to production-scale models with non-standard architectures (Mixture-of-Experts routing, linear recurrence layers, hybrid attention) remains an open question.

We address this question by training SAEs on Qwen~3.5-35B-A3B \citep{qwen2025qwen3}, a 35B-parameter MoE model (3B active per token) that uses a hybrid architecture: repeating blocks of three GatedDeltaNet layers \citep{yang2024gated} followed by one standard attention layer.
We target five \emph{agentic} behavioral traits (autonomy, tool-use eagerness, persistence, risk calibration, and deference) that are central to the deployment characteristics of AI agents.

Our method bridges SAE-based feature discovery with activation steering.
For each trait, we:
(1)~collect contrastive activation pairs from the model's residual stream,
(2)~train a linear probe on the SAE's latent activations to predict the trait,
(3)~project the probe weights through the SAE's decoder to obtain a steering vector in the model's native 2048-dimensional activation space.
This projection bypasses the SAE's top-$k$ activation function, which would otherwise discretize the latent space and destroy fine-grained directional information.

We evaluate each steering vector by running 50 ReAct-style agent scenarios and measuring behavioral change through proxy metrics extracted from raw trajectories.
Our contributions are:

\begin{enumerate}
    \item \textbf{SAE-decoded probe steering at 35B MoE scale.} We demonstrate that the method $\mathbf{v} = W_{\text{dec}}^\top \mathbf{w}_{\text{probe}}$ produces causally effective steering vectors on a production-scale hybrid architecture, achieving $d = 1.01$ for autonomy.

    \item \textbf{A dominant agency axis.} Cross-trait analysis reveals that all five steering vectors primarily modulate a single shared disposition (acting independently versus deferring), with trait-specific effects appearing only in tool-type composition and dose-response dynamics.

    \item \textbf{A correlation/causation dissociation.} Risk calibration and tool-use eagerness, probed from the \emph{same} SAE with nearly identical $R^2$, yield orthogonal vectors with opposite causal efficacy. High probe accuracy does not guarantee causal relevance.

    \item \textbf{Prefill-only behavioral commitment.} Steering during autoregressive decoding has zero effect. Behavioral decisions are committed during the prefill phase and propagated through the GatedDeltaNet recurrence.
\end{enumerate}

%=============================================================================
\section{Related Work}
\label{sec:related}
%=============================================================================

\paragraph{Sparse Autoencoders for Interpretability.}
SAEs decompose model activations into sparse, interpretable features by training an overcomplete autoencoder with a sparsity constraint \citep{cunningham2023sparse, bricken2023monosemanticity}.
\citet{templeton2024scaling} scaled SAEs to Claude~3 Sonnet and demonstrated that individual features correspond to human-interpretable concepts.
\citet{gao2024scaling} introduced the top-$k$ activation function and auxiliary dead-feature losses, improving training stability at scale.
Our work extends SAE training to hybrid GatedDeltaNet/attention architectures and uses SAE latents as an intermediate representation for steering vector construction rather than directly clamping features.

\paragraph{Activation Steering.}
\citet{turner2023activation} introduced activation addition, where a direction vector is added to the residual stream during the forward pass to shift model behavior.
\citet{li2023inference} used probing to identify attention heads encoding truthfulness, then shifted activations at those heads during inference.
\citet{rimsky2024steering} used contrastive activation addition (CAA) to steer model behavior on Llama~2, computing steering vectors as the mean difference between contrastive prompt pairs.
Our SAE-decoded probe method differs from mean-diff CAA by leveraging the SAE's learned basis to identify more targeted directions; a head-to-head comparison is reported as a limitation (\S\ref{sec:limitations}).

\paragraph{Linear Probing and Causal Intervention.}
Linear probes have been widely used to test whether specific information is linearly represented in model activations \citep{belinkov2022probing, burns2023discovering}.
However, probes measure \emph{correlation} between representations and labels, not causal influence on model behavior.
\citet{ravfogel2020null} and \citet{elazar2021amnesiac} explored the gap between probe accuracy and causal relevance through nullspace projection and amnesic counterfactuals, respectively.
Our risk-calibration dissociation provides an empirical case study of this gap: a probe with $R^2 = 0.795$ that identifies a direction with zero causal efficacy for steering, while a probe with $R^2 = 0.792$ on the same SAE successfully steers a different trait.

\paragraph{Hybrid and Linear-Recurrence Architectures.}
GatedDeltaNet \citep{yang2024gated} combines gated linear attention with delta-rule updates, offering sub-quadratic sequence processing.
Qwen~3.5 \citep{qwen2025qwen3} deploys this in a hybrid configuration alternating recurrence and attention layers.
The interaction between linear recurrence and sparse feature decomposition has not been previously studied.
Our finding that behavioral commitments are computed during prefill and propagated through the recurrent state provides evidence about the computational role of these architectural components.

%=============================================================================
\section{Model and Architecture}
\label{sec:model}
%=============================================================================

We study Qwen~3.5-35B-A3B, a Mixture-of-Experts language model with 35 billion total parameters and approximately 3 billion active per token.
The model has 40 transformer layers organized into 10 blocks of 4 layers each: 3 GatedDeltaNet layers followed by 1 standard multi-head attention layer (Table~\ref{tab:architecture}).
Each MoE layer routes tokens to 8 of 256 experts plus 1 shared expert, with per-expert intermediate dimension 512.
The residual stream has hidden dimension $d = 2048$.

\begin{table}[h]
\centering
\caption{Qwen~3.5-35B-A3B architecture summary.}
\label{tab:architecture}
\small
\begin{tabular}{ll}
\toprule
\textbf{Property} & \textbf{Value} \\
\midrule
Total parameters & 35B \\
Active parameters/token & $\sim$3B \\
Layers & 40 (10 blocks $\times$ 4) \\
Block pattern & 3 GatedDeltaNet + 1 Attention \\
Hidden dimension & 2,048 \\
MoE experts & 256 total, 8 routed + 1 shared \\
Expert intermediate dim & 512 \\
\bottomrule
\end{tabular}
\end{table}

The hybrid architecture creates a natural experimental contrast: within each block, we can compare feature representations in GatedDeltaNet (recurrent) versus attention (quadratic) layers at matched depth.

%=============================================================================
\section{Method}
\label{sec:method}
%=============================================================================

\begin{figure}[t]
\centering
\includegraphics[width=\textwidth]{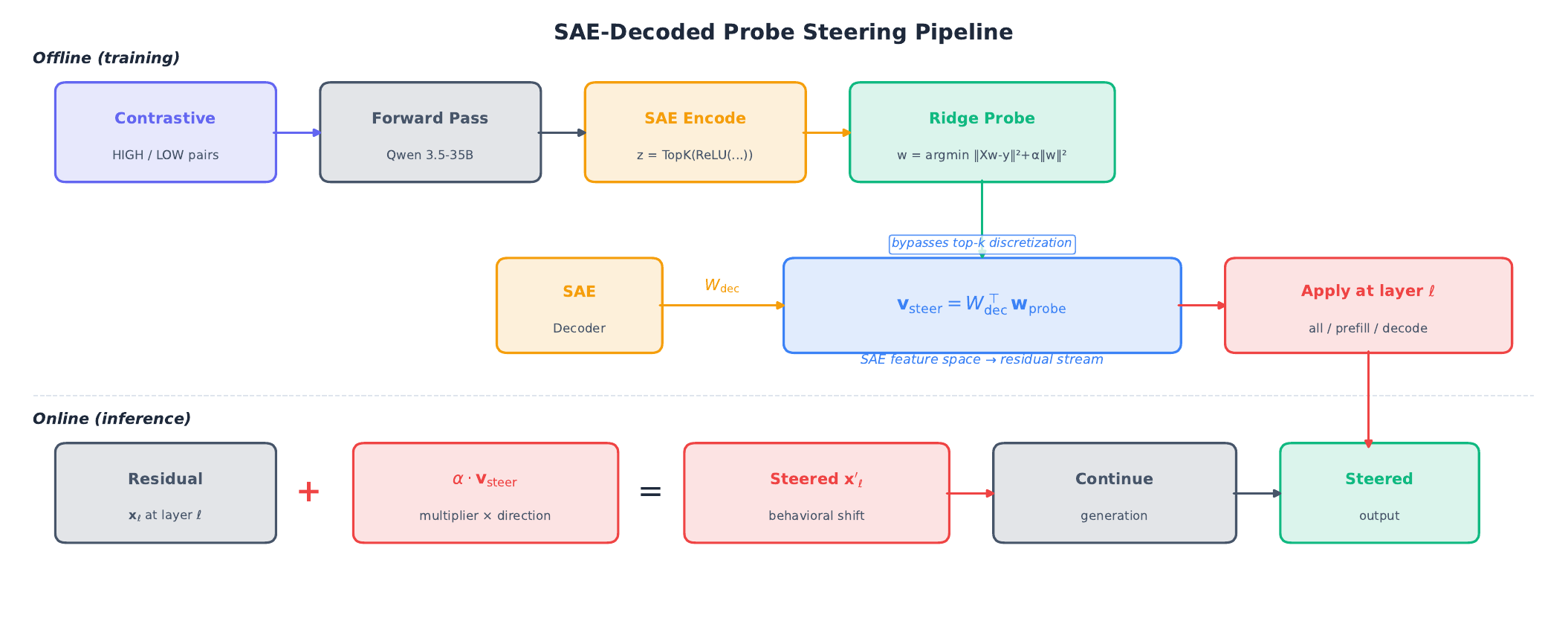}
\caption{SAE-decoded probe steering pipeline. Contrastive activations are encoded through the SAE, a ridge regression probe identifies the discriminative direction in SAE latent space, and the probe weights are projected through the decoder to obtain a continuous steering vector in the model's native activation space.}
\label{fig:pipeline}
\end{figure}

Our pipeline has four stages: SAE training (\S\ref{sec:sae_training}), contrastive feature identification (\S\ref{sec:contrastive}), probe-to-steering-vector projection (\S\ref{sec:probe_projection}), and behavioral evaluation (\S\ref{sec:evaluation}).
Figure~\ref{fig:pipeline} summarizes the full pipeline.

%-----------------------------------------------------------------------------
\subsection{SAE Training}
\label{sec:sae_training}
%-----------------------------------------------------------------------------

We train 9 top-$k$ SAEs on residual stream activations at four depth levels (early, early-mid, mid, late), covering both GatedDeltaNet and attention sublayers at each depth (Table~\ref{tab:saes}).
An additional control SAE at a different position within the mid block tests within-block position effects.

\begin{table}[h]
\centering
\caption{SAE configurations. All SAEs use top-$k$ activation with auxiliary dead-feature loss. Learning rates are scaled inversely with activation norm at each depth.}
\label{tab:saes}
\small
\begin{tabular}{lcccccc}
\toprule
\textbf{SAE ID} & \textbf{Layer} & \textbf{Type} & \textbf{Block} & \textbf{Dict Size} & $k$ & \textbf{LR} \\
\midrule
\texttt{sae\_delta\_early}     & 6  & DeltaNet  & 1 & 8{,}192  & 128 & 3e-5 \\
\texttt{sae\_attn\_early}      & 7  & Attention & 1 & 8{,}192  & 128 & 3e-5 \\
\texttt{sae\_delta\_earlymid}  & 14 & DeltaNet  & 3 & 16{,}384 & 96  & 2e-5 \\
\texttt{sae\_attn\_earlymid}   & 15 & Attention & 3 & 16{,}384 & 96  & 2e-5 \\
\texttt{sae\_delta\_mid\_pos1} & 21 & DeltaNet  & 5 & 16{,}384 & 64  & 1e-5 \\
\texttt{sae\_delta\_mid}       & 22 & DeltaNet  & 5 & 16{,}384 & 64  & 1e-5 \\
\texttt{sae\_attn\_mid}        & 23 & Attention & 5 & 16{,}384 & 64  & 1e-5 \\
\texttt{sae\_delta\_late}      & 34 & DeltaNet  & 8 & 16{,}384 & 64  & 8e-6 \\
\texttt{sae\_attn\_late}       & 35 & Attention & 8 & 16{,}384 & 64  & 8e-6 \\
\bottomrule
\end{tabular}
\end{table}

Each SAE is trained on 200M tokens from a mixture of HuggingFaceH4/ultrachat\_200k, allenai/WildChat-1M (GDPR-filtered), and synthetic tool-use conversations.
We follow the FAST methodology \citep{gao2024scaling}: sequential conversation processing with a 2M-vector circular CPU buffer for activation mixing.
Dead features are resampled every 5{,}000 steps using an auxiliary top-$k$ loss on the reconstruction residual \citep{gao2024scaling}, with learning rate warmup over 1{,}000 steps and linear decay over the final 20\% of training.

The SAE architecture follows standard practice.
Given a residual stream activation $\mathbf{x} \in \mathbb{R}^d$, the SAE computes:
\begin{align}
\mathbf{z} &= \text{TopK}\big(\text{ReLU}(W_{\text{enc}}(\mathbf{x} - \mathbf{b}_{\text{pre}}))\big) \label{eq:encode} \\
\hat{\mathbf{x}} &= W_{\text{dec}}^\top \mathbf{z} + \mathbf{b}_{\text{pre}} \label{eq:decode}
\end{align}
where $W_{\text{enc}} \in \mathbb{R}^{D \times d}$, $W_{\text{dec}} \in \mathbb{R}^{D \times d}$ with columns normalized to unit norm, $\mathbf{b}_{\text{pre}} \in \mathbb{R}^d$ is a learned centering parameter, $D$ is the dictionary size, and $\text{TopK}$ retains only the $k$ largest activations, zeroing the rest.

%-----------------------------------------------------------------------------
\subsection{Contrastive Feature Identification}
\label{sec:contrastive}
%-----------------------------------------------------------------------------

We define five agentic behavioral traits, each decomposed into three sub-behaviors (Table~\ref{tab:traits}).

\begin{table}[h]
\centering
\caption{Behavioral traits and sub-behaviors.}
\label{tab:traits}
\small
\begin{tabular}{lp{8cm}}
\toprule
\textbf{Trait} & \textbf{Sub-behaviors} \\
\midrule
Autonomy & decision independence, action initiation, permission avoidance \\
Tool-use eagerness & tool reach, proactive info gathering, tool diversity \\
Persistence & retry willingness, strategy variation, escalation reluctance \\
Risk calibration & approach novelty, scope expansion, uncertainty tolerance \\
Deference & instruction literalness, challenge avoidance, suggestion restraint \\
\bottomrule
\end{tabular}
\end{table}

For each trait, we construct contrastive prompt pairs: HIGH variants that should elicit strong expression of the trait and LOW variants that should suppress it.
We generate 800 composite pairs (10 templates $\times$ 4 variations $\times$ 4 task domains $\times$ 5 traits) plus 720 sub-behavior-specific pairs, for 1{,}520 total.
Task domains (coding, research, communication, data analysis) ensure breadth.

For each pair, we run the model's forward pass and extract residual stream activations at all 9 SAE hook points.
We encode these through each SAE and compute the \textbf{Trait Association Score} (TAS) for each latent feature $j$:
\begin{equation}
    \text{TAS}_j = \frac{\mathbb{E}[z_j^{\text{high}}] - \mathbb{E}[z_j^{\text{low}}]}{\text{std}(z_j^{\text{high}} - z_j^{\text{low}})}
    \label{eq:tas}
\end{equation}
where $z_j^{\text{high}}$ and $z_j^{\text{low}}$ are feature $j$'s activations on HIGH and LOW prompts respectively, pooled at the last token position to avoid sequence-length confounds.
We select the SAE with the highest mean $|\text{TAS}|$ for each trait as the best hook point for subsequent probing.

%-----------------------------------------------------------------------------
\subsection{Probe-to-Residual-Stream Projection}
\label{sec:probe_projection}
%-----------------------------------------------------------------------------

We construct steering vectors in two steps, translating from SAE feature space to the model's native activation space.

\paragraph{Step 1: Linear Probe.}
For each trait, we train a ridge regression probe (L2-regularized linear regression) on the SAE latent activations from the best SAE, with binary labels $y = 1$ for HIGH and $y = 0$ for LOW:
\begin{equation}
    \hat{y} = \mathbf{w}_{\text{probe}}^\top \mathbf{z} + b, \quad \mathbf{w}_{\text{probe}} = \arg\min_{\mathbf{w}} \| X\mathbf{w} - \mathbf{y} \|^2 + \alpha \| \mathbf{w} \|^2
\end{equation}
where $\mathbf{z} \in \mathbb{R}^D$ is the SAE encoding of a contrastive example, $\mathbf{w}_{\text{probe}} \in \mathbb{R}^D$ are the learned weights, and $\alpha$ is selected from $\{0.01, 0.1, 1.0, 10.0\}$ by held-out $R^2$.
The probe's $R^2$ quantifies how linearly separable the trait is in SAE latent space.

\paragraph{Step 2: Decoder Projection.}
We project the probe weights through the SAE decoder to obtain a steering vector in the residual stream:
\begin{equation}
    \mathbf{v}_{\text{steer}} = W_{\text{dec}}^\top \mathbf{w}_{\text{probe}}
    \label{eq:steering_vec}
\end{equation}
where $W_{\text{dec}} \in \mathbb{R}^{D \times d}$ is the SAE's decoder weight matrix.
This maps the discriminative direction in $D$-dimensional SAE space to a $d$-dimensional direction in the residual stream.

\paragraph{Why not steer inside the SAE?}
The SAE's top-$k$ activation function (Eq.~\ref{eq:encode}) introduces a hard threshold: only $k$ features are nonzero per token.
Amplifying a feature that falls outside the top-$k$ has no effect on the reconstruction, and modifying a feature that is inside the top-$k$ shifts the reconstruction in a quantized, $k$-sparse manner.
The decoder projection circumvents this by operating in continuous activation space. The resulting vector $\mathbf{v}_{\text{steer}}$ can be added at any scale without discretization artifacts.

\paragraph{Steering Application.}
During inference, we add the steering vector to the residual stream at the probe's layer:
\begin{equation}
    \mathbf{x}'_\ell = \mathbf{x}_\ell + \alpha \cdot \mathbf{v}_{\text{steer}}
\end{equation}
where $\alpha$ is a scalar multiplier.
We test three application modes: \texttt{all\_positions} (every token in both prefill and decode), \texttt{prefill\_only} (only during prompt processing), and \texttt{decode\_only} (only during autoregressive generation).

%-----------------------------------------------------------------------------
\subsection{Behavioral Evaluation}
\label{sec:evaluation}
%-----------------------------------------------------------------------------

We evaluate steering effects using 50 ReAct-style agent scenarios spanning four task domains.
Each scenario provides the model with tool schemas (\texttt{code\_execute}, \texttt{web\_search}, \texttt{file\_read}, \texttt{file\_write}, \texttt{ask\_user}) and pre-cached tool responses to eliminate network dependencies.
The model generates up to 5 turns of tool calls and responses.

We extract behavioral proxy metrics directly from trajectories:
\begin{itemize}
    \item \textbf{Autonomy proxy}: 1 if no \texttt{ask\_user} calls, else 0
    \item \textbf{Tool-use proxy}: count of non-\texttt{ask\_user} tool calls
    \item \textbf{Persistence proxy}: number of agent turns
    \item \textbf{Risk proxy}: fraction of tool calls that are \texttt{code\_execute} or \texttt{file\_write}
    \item \textbf{Deference proxy}: 1 if any \texttt{ask\_user} call, else 0
\end{itemize}

We use \texttt{ask\_user} count and proactive tool count (non-\texttt{ask\_user}) as primary metrics, reporting Cohen's $d$ (standardized effect size) and Mann-Whitney $U$ tests (non-parametric, appropriate for count data).
Statistical significance is assessed with Bonferroni correction for 35 total conditions ($\alpha_{\text{corrected}} = 0.0014$).

%=============================================================================
\section{Results}
\label{sec:results}
%=============================================================================

We ran 1{,}800 agent rollouts: 50 scenarios $\times$ (1 baseline + 35 steered conditions).
The unsteered model predominantly asks the user for help (78\% of tool calls are \texttt{ask\_user}), with 38\% of scenarios producing zero tool calls.

%-----------------------------------------------------------------------------
\subsection{Per-Trait Steering Efficacy}
\label{sec:per_trait}
%-----------------------------------------------------------------------------

Table~\ref{tab:main_results} summarizes the best steering condition for each trait.

\begin{table}[h]
\centering
\caption{Best steering condition per trait. $d(\text{ask})$: Cohen's $d$ for \texttt{ask\_user} reduction. $d(\text{pro})$: Cohen's $d$ for proactive tool-call increase. $\emptyset$TC\%: fraction of scenarios with zero tool calls (coherence proxy). Conditions tested: multipliers $\alpha \in \{1, 2, 3\}$; positions $\in \{\text{all, prefill, decode}\}$.}
\label{tab:main_results}
\small
\begin{tabular}{lcccccccc}
\toprule
\textbf{Trait} & \textbf{SAE} & \textbf{Layer} & $R^2$ & \textbf{Condition} & $d(\text{ask})$ & $d(\text{pro})$ & $\emptyset$TC\% & \textbf{Tier} \\
\midrule
Autonomy & \texttt{delta\_mid} & 22 & 0.865 & all, $\alpha\!=\!3$ & $-$0.87 & \textcolor{tierone}{\textbf{+1.04}} & 36\% & \textcolor{tierone}{1} \\
Tool use & \texttt{delta\_mid\_p1} & 21 & 0.792 & all, $\alpha\!=\!3$ & $-$0.60 & \textcolor{tiertwo}{\textbf{+0.39}} & 26\% & \textcolor{tiertwo}{2} \\
Deference & \texttt{attn\_earlymid} & 15 & 0.737 & pre, $\alpha\!=\!2$ & $-$0.58 & \textcolor{tiertwo}{\textbf{+0.80}} & 46\% & \textcolor{tiertwo}{3} \\
Persistence & \texttt{attn\_mid} & 23 & 0.561 & all, $\alpha\!=\!1$ & $-$0.45 & +0.28 & 40\% & \textcolor{tierfail}{fail} \\
Risk cal. & \texttt{delta\_mid\_p1} & 21 & 0.795 & --- & $-$0.51 & +0.00 & 70\% & \textcolor{tierfail}{fail} \\
\bottomrule
\end{tabular}
\end{table}

\paragraph{Autonomy (Tier~1).}
At $\alpha = 2.0$ (all positions), \texttt{ask\_user} calls drop from $1.04 \pm 1.65$ to $0.12 \pm 0.33$ ($d = -0.76$, $p < 0.0001$), while proactive tool calls rise from $0.30 \pm 0.85$ to $2.10 \pm 2.14$ ($d = +1.01$, $p < 0.0001$; 95\% CI $[+0.60, +1.43]$).
At $\alpha = 3.0$, the effect on proactive calls is maintained ($d = +1.04$) with slightly more coherence degradation ($\emptyset$TC = 36\% vs.\ 30\%).
Above $\alpha = 5.0$, the model collapses entirely ($\emptyset$TC = 100\%).
The dose-response follows a smooth inverted-U curve.

\paragraph{Tool-Use Eagerness (Tier~2).}
This trait shows a phase-transition dose-response.
At $\alpha = 2.0$, 58\% of scenarios have zero tool calls, paradoxically \emph{worse} than baseline (38\%).
At $\alpha = 3.0$, the model passes through a formatting instability into a new behavioral mode: 4.26 tool calls per scenario (vs.\ 1.34 baseline), 88\% of which are \texttt{web\_search}.
The perturbation at $\alpha = 3.0$ corresponds to 12.8$\times$ the residual stream RMS norm, yet the model maintains coherent output in 74\% of scenarios.

\paragraph{Deference (Tier~3, borderline).}
The only trait where \texttt{prefill\_only} outperforms \texttt{all\_positions} ($d(\text{pro}) = +0.80$ vs.\ $+0.76$).
This suggests deference is a dispositional property computed during prompt processing.
The therapeutic window is narrow: in prefill-only mode, $\alpha = 1.0$ has a weak effect ($d(\text{pro}) = +0.23$), $\alpha = 2.0$ produces genuine redirection ($d(\text{pro}) = +0.80$), and $\alpha = 3.0$ degrades coherence ($\emptyset$TC = 84\%).
In all-positions mode, $\alpha = 3.0$ collapses output entirely ($\emptyset$TC = 100\%).
Deference operates through an attention SAE (layer~15, early-mid block), the earliest intervention point and the only attention-layer SAE that produces genuine behavioral redirection.

\paragraph{Persistence (Fail).}
The weakest probe ($R^2 = 0.561$) produces the weakest steering.
No multiplier achieves significant behavioral redirection; the best condition ($\alpha = 1.0$, all positions: $d(\text{ask}) = -0.45$, $p = 0.056$) does not approach the Bonferroni-corrected threshold ($\alpha_{\text{corrected}} = 0.0014$).
All higher multipliers produce pure suppression without compensatory proactive behavior.

\paragraph{Risk Calibration (Fail).}
Discussed in detail in \S\ref{sec:dissociation}.

%-----------------------------------------------------------------------------
\subsection{Dose-Response Taxonomy}
\label{sec:dose_response}
%-----------------------------------------------------------------------------

\begin{figure}[t]
\centering
\includegraphics[width=\textwidth]{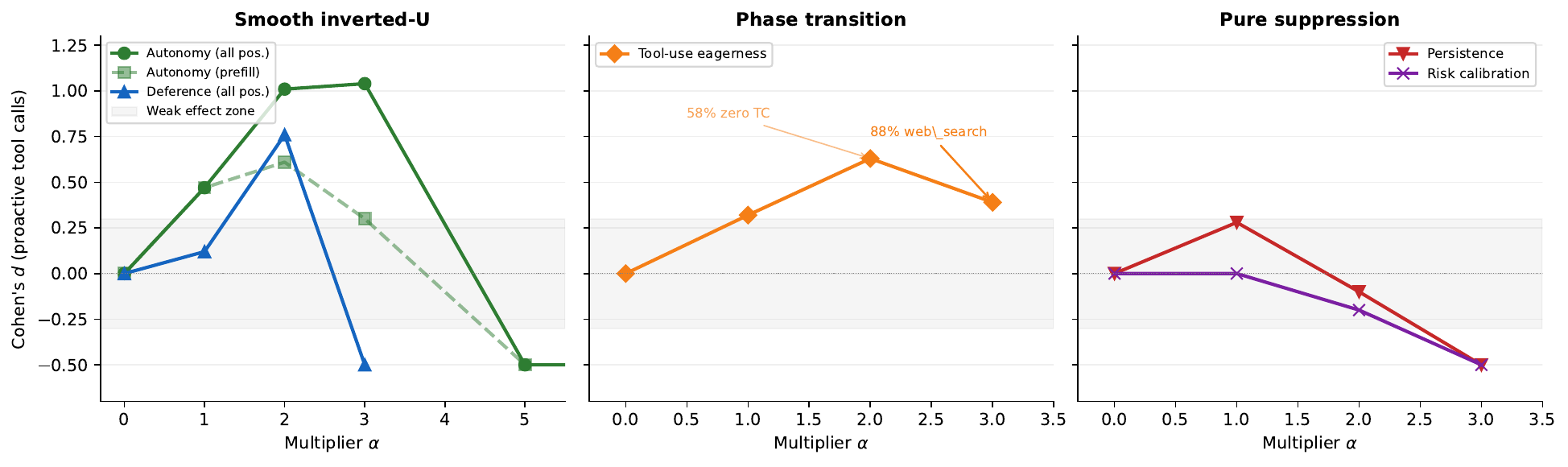}
\caption{Dose-response curves for proactive tool-call effect size $d(\text{pro})$ as a function of steering multiplier $\alpha$. Autonomy exhibits a smooth inverted-U with a clear therapeutic window. Tool-use eagerness shows a phase transition: intermediate multipliers degrade performance before a high-$\alpha$ regime produces a new behavioral mode. Persistence and risk calibration show monotonic suppression at all multipliers.}
\label{fig:dose_response}
\end{figure}

The five traits exhibit three qualitatively distinct dose-response patterns (Figure~\ref{fig:dose_response}):

\begin{enumerate}
    \item \textbf{Smooth inverted-U} (autonomy, deference): Effect grows monotonically with $\alpha$ until a sharp cliff. There exists a clear therapeutic window between weak ($d < 0.3$) and collapsed ($\emptyset$TC $> 80\%$) regimes.

    \item \textbf{Phase transition} (tool-use eagerness): Non-monotonic. Intermediate multipliers \emph{degrade} performance; high multipliers push through a formatting instability into a new stable behavioral attractor with qualitatively different output.

    \item \textbf{Pure suppression} (persistence, risk calibration): Monotonic degradation. Increasing $\alpha$ suppresses the dominant behavior (\texttt{ask\_user}) without replacing it with alternative actions. No multiplier redirects behavior.
\end{enumerate}

These patterns are not predicted by probe $R^2$, vector norm, SAE type, or layer depth (Table~\ref{tab:main_results}).
Risk calibration ($R^2 = 0.795$) fails while deference ($R^2 = 0.737$) succeeds.
The autonomy vector (norm 10.58) tolerates $\alpha = 3.0$, while the deference vector (norm 28.54, perturbation 10.2$\times$ RMS) collapses at $\alpha = 3.0$.
Coherence under perturbation depends on the \emph{direction}, not just the \emph{magnitude}, of the intervention.

%-----------------------------------------------------------------------------
\subsection{The Decode-Only Null Result}
\label{sec:decode_only}
%-----------------------------------------------------------------------------

We tested autonomy steering (the strongest trait) in \texttt{decode\_only} mode, where the steering vector is applied only during autoregressive token generation, not during prompt processing.
At both $\alpha = 1.0$ and $\alpha = 2.0$, the effect is null:

\begin{table}[h]
\centering
\caption{Decode-only steering produces zero effect on autonomy.}
\label{tab:decode_only}
\small
\begin{tabular}{lccccc}
\toprule
\textbf{Condition} & $d(\text{ask})$ & $p$ & $d(\text{pro})$ & $p$ & $\emptyset$TC\% \\
\midrule
decode-only $\alpha = 1.0$ & $-$0.16 & 0.674 & $+$0.13 & 0.342 & 38\% \\
decode-only $\alpha = 2.0$ & $-$0.20 & 0.359 & $+$0.02 & 0.511 & 38\% \\
\midrule
all-positions $\alpha = 2.0$ & $-$0.76 & $<$0.0001 & $+$1.01 & $<$0.0001 & 30\% \\
prefill-only $\alpha = 2.0$ & $-$0.64 & 0.0007 & $+$0.61 & 0.002 & 42\% \\
\bottomrule
\end{tabular}
\end{table}

This provides causal evidence that the model computes its behavioral commitments (whether to ask or act) during the prefill phase and propagates them through the GatedDeltaNet recurrent state.
By the time the model is generating tokens, the decision is already embedded in the hidden state.
The attention-based steering for deference corroborates this: \texttt{prefill\_only} actually \emph{outperforms} \texttt{all\_positions} for that trait, consistent with deference being a dispositional property set during prompt processing.

%-----------------------------------------------------------------------------
\subsection{Tool-Type Composition}
\label{sec:tool_types}
%-----------------------------------------------------------------------------

Even where the cross-trait specificity matrix shows shared agency effects (\S\ref{sec:cross_trait}), the \emph{type} of action differs qualitatively between steering vectors:

\begin{table}[h]
\centering
\caption{Tool-type breakdown for the three successful steering conditions vs.\ baseline.}
\label{tab:tool_types}
\small
\begin{tabular}{lccccc}
\toprule
\textbf{Condition} & \textbf{ask\_user} & \textbf{web\_search} & \textbf{code\_exec} & \textbf{file\_r/w} & \textbf{Total} \\
\midrule
Baseline & 78\% & 22\% & 0\% & 0\% & 67 \\
Autonomy ($\alpha\!=\!2$) & 5\% & 44\% & 48\% & 3\% & 111 \\
Tool use ($\alpha\!=\!3$) & 6\% & 88\% & 5\% & $<$1\% & 213 \\
Deference ($\alpha\!=\!2$, pre) & 16\% & 63\% & 15\% & 6\% & 87 \\
\bottomrule
\end{tabular}
\end{table}

The autonomy vector unlocks code execution (0\% $\to$ 48\% of tool calls), the tool-use vector creates compulsive web searching (88\%), and the deference vector produces a balanced portfolio.
These represent qualitatively different behavioral modes even though all three increase the overall disposition to act.

%-----------------------------------------------------------------------------
\subsection{Cross-Trait Specificity}
\label{sec:cross_trait}
%-----------------------------------------------------------------------------

\begin{figure}[t]
\centering
\includegraphics[width=0.65\textwidth]{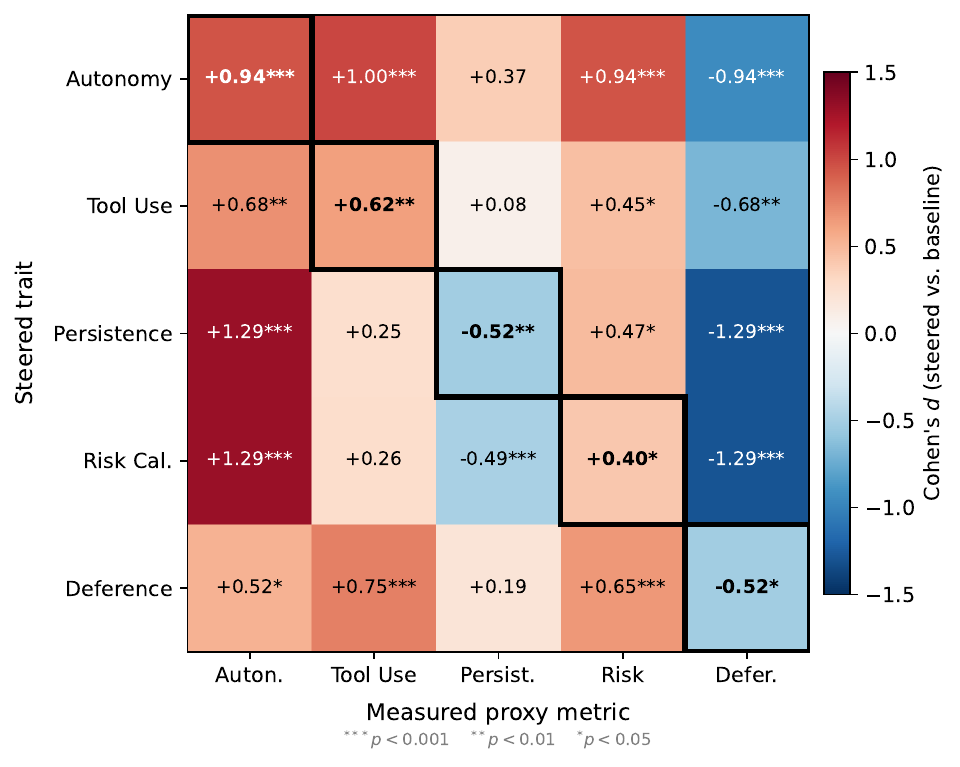}
\caption{Cross-trait specificity matrix (Cohen's $d$, $\alpha = 2.0$, all positions). Every steering vector primarily increases autonomy and decreases deference, revealing a dominant agency axis. No vector achieves a specificity ratio $> 1.0$. Numeric values in Table~\ref{tab:cross_trait}.}
\label{fig:cross_trait}
\end{figure}

We measured each steering vector's effect on all five proxy metrics to assess trait specificity (Table~\ref{tab:cross_trait}, Figure~\ref{fig:cross_trait}).

\begin{table}[h]
\centering
\caption{Cross-trait specificity matrix. Each cell is Cohen's $d$ (steered vs.\ baseline at $\alpha = 2.0$, all positions). Bold diagonal = on-target effect. $^{***}$ $p < 0.001$, $^{**}$ $p < 0.01$, $^{*}$ $p < 0.05$.}
\label{tab:cross_trait}
\small
\begin{tabular}{l|ccccc}
\toprule
\textbf{Steered} & \textbf{Auton.} & \textbf{Tool Use} & \textbf{Persist.} & \textbf{Risk} & \textbf{Defer.} \\
\midrule
Autonomy & \textbf{0.94}$^{***}$ & 1.00$^{***}$ & 0.37 & 0.94$^{***}$ & $-$0.94$^{***}$ \\
Tool use & 0.68$^{**}$ & \textbf{0.62}$^{**}$ & 0.08 & 0.45$^{*}$ & $-$0.68$^{**}$ \\
Persistence & 1.29$^{***}$ & 0.25 & \textbf{$-$0.52}$^{**}$ & 0.47$^{*}$ & $-$1.29$^{***}$ \\
Risk cal. & 1.29$^{***}$ & 0.26 & $-$0.49$^{***}$ & \textbf{0.40}$^{*}$ & $-$1.29$^{***}$ \\
Deference & 0.52$^{*}$ & 0.75$^{***}$ & 0.19 & 0.65$^{***}$ & \textbf{$-$0.52}$^{*}$ \\
\bottomrule
\end{tabular}
\end{table}

\paragraph{Finding: A dominant agency axis.}
Every steering vector primarily increases autonomy and decreases deference, regardless of which trait it was trained to target.
No vector achieves a specificity ratio (on-target $|d|$ / max off-target $|d|$) greater than 1.0.
The autonomy vector's strongest effect is actually on the \emph{tool-use} proxy ($d = 1.00$), exceeding its on-target autonomy effect ($d = 0.94$).
The persistence and risk-calibration vectors produce their \emph{largest} effects on the autonomy proxy ($d = 1.29$), substantially exceeding their on-target effects ($d = 0.52$ and $d = 0.40$).

This is not simply measurement artifact from correlated proxies.
The autonomy and deference proxies are definitionally anti-correlated (both derived from \texttt{ask\_user}), but the tool-use and risk proxies are constructed from independent counts.
All five vectors modulate a shared \emph{agency} disposition, with trait-specific effects appearing as secondary modulations in tool-type composition (\S\ref{sec:tool_types}) and dose-response dynamics (\S\ref{sec:dose_response}).

%-----------------------------------------------------------------------------
\subsection{The Risk-Calibration Dissociation}
\label{sec:dissociation}
%-----------------------------------------------------------------------------

Risk calibration and tool-use eagerness share the same SAE (\texttt{sae\_delta\_mid\_pos1}, layer~21) with nearly identical probe accuracy ($R^2 = 0.795$ vs.\ $0.792$) and vector norms (22.45 vs.\ 23.84).
Yet the tool-use vector steers behavior ($d(\text{pro}) = +0.39$) while the risk-calibration vector produces only suppression ($d(\text{pro}) = 0.00$ at best).

The probe directions are nearly orthogonal:

\begin{table}[h]
\centering
\caption{Geometry of the risk-calibration and tool-use probe directions.}
\label{tab:dissociation}
\small
\begin{tabular}{lc}
\toprule
\textbf{Metric} & \textbf{Value} \\
\midrule
Cosine similarity (SAE feature space) & $-$0.017 \\
Cosine similarity (residual stream) & $-$0.065 \\
Risk-cal variance parallel to tool-use & 0.03\% \\
Risk-cal variance orthogonal to tool-use & 99.97\% \\
\bottomrule
\end{tabular}
\end{table}

The risk-calibration probe found a direction that is:
\begin{itemize}
    \item \textbf{Predictive}: $R^2 = 0.795$ on held-out contrastive pairs
    \item \textbf{Distinct}: nearly orthogonal to the tool-use direction (cosine $= -0.017$)
    \item \textbf{Concentrated}: 83 features explain 50\% of probe variance (vs.\ 113 for tool use)
    \item \textbf{Causally inert}: produces only suppression, no behavioral redirection
\end{itemize}

This is a controlled dissociation between predictive and causal relevance.
The risk-calibration direction is a real statistical pattern in the model's activations, an \emph{epiphenomenal representation} that co-occurs with cautious/bold behavior but is not part of the causal chain generating it.
The failure cannot be attributed to shared variance with tool use (orthogonal directions), weak probing (high $R^2$), or wrong layer (same SAE and layer as a successful vector).

\textbf{Implication}: Probe $R^2$ measures correlation between the learned direction and the trait label distribution.
It does not guarantee that the direction is causally upstream of the behavior it predicts.
Intervention experiments are necessary to distinguish causal from epiphenomenal representations.

%-----------------------------------------------------------------------------
\subsection{Feature Attribution}
\label{sec:feature_attribution}
%-----------------------------------------------------------------------------

We decompose each steering vector by its constituent SAE features (Table~\ref{tab:concentration}).

\begin{table}[h]
\centering
\caption{Feature concentration of steering vectors. Values indicate how many SAE features (out of 16{,}384) are needed to explain a given fraction of the vector's squared norm.}
\label{tab:concentration}
\small
\begin{tabular}{lcccc}
\toprule
\textbf{Trait} & \textbf{50\% var} & \textbf{80\% var} & \textbf{90\% var} & \textbf{95\% var} \\
\midrule
Autonomy & 48 (0.3\%) & 147 (0.9\%) & 228 (1.4\%) & 315 (1.9\%) \\
Tool use & 113 (0.7\%) & 316 (1.9\%) & 462 (2.8\%) & 595 (3.6\%) \\
Persistence & 77 (0.5\%) & 227 (1.4\%) & 339 (2.1\%) & 439 (2.7\%) \\
Risk cal. & 83 (0.5\%) & 257 (1.6\%) & 403 (2.5\%) & 529 (3.2\%) \\
Deference & 70 (0.4\%) & 216 (1.3\%) & 340 (2.1\%) & 453 (2.8\%) \\
\bottomrule
\end{tabular}
\end{table}

All directions are extremely sparse: $<$1\% of features explain 50\% of the steering signal.
Autonomy is the most concentrated (48 features for 50\%), which may explain its efficacy: a concentrated signal may be harder for the model to ``route around.''

Despite the behavioral cross-talk observed in \S\ref{sec:cross_trait}, the steering vectors are nearly orthogonal in residual stream space (max pairwise cosine similarity = 0.086 between autonomy and risk calibration).
The shared agency axis in \emph{behavior space} is therefore not a single direction in \emph{activation space}.
Multiple distinct perturbations converge to the same behavioral attractor.

%=============================================================================
\section{Discussion}
\label{sec:discussion}
%=============================================================================

\paragraph{One Axis, Not Five.}
We set out to find five independent behavioral directions and instead found a dominant agency axis with secondary trait-specific modulations.
Agentic behavior in this model appears to be organized around a single ask-or-act decision point, with the type of action (code execution vs.\ web search vs.\ balanced portfolio) as a secondary consideration.

The structure is analogous to findings in personality psychology, where the ``Big Five'' traits are empirically correlated and factor analyses sometimes extract a ``General Factor of Personality'' \citep{musek2007general} that subsumes the individual traits.
Whether this reflects a genuine architectural constraint (a single ``agency bottleneck'' in the forward pass) or an artifact of our proxy metrics (which are all derived from tool-call patterns) is an open question that requires richer evaluation metrics to resolve.

\paragraph{The Probe/Causation Gap.}
The risk-calibration dissociation provides an empirical demonstration that high linear probe accuracy does not imply causal relevance for steering.
Prior work has argued this point on theoretical grounds \citep{ravfogel2020null, belinkov2022probing}. Our case is unusually controlled: same SAE, same layer, same $R^2$, orthogonal directions, opposite causal efficacy.
We recommend that future work using probes for steering report intervention results alongside $R^2$, treating probe accuracy as a necessary but insufficient condition.

\paragraph{Prefill as the Behavioral Commitment Phase.}
The decode-only null result has implications for both interpretability and deployment.
For interpretability, it suggests that GatedDeltaNet recurrent layers ``compile'' behavioral decisions during prefill into the recurrent state, which then deterministically guides generation.
For deployment, it implies that monitoring or intervening on the prefill phase may be more effective than monitoring generated tokens for behavioral anomalies.

\paragraph{Therapeutic Windows Are Trait-Specific.}
There is no universal multiplier that works across traits.
Autonomy tolerates $\alpha = 3.0$. Deference has a narrow window at exactly $\alpha = 2.0$. Tool use requires a phase-transition-inducing $\alpha = 3.0$.
Production deployment of behavioral steering would require per-trait calibration, not a single scaling parameter.

%=============================================================================
\section{Limitations and Future Work}
\label{sec:limitations}
%=============================================================================

\paragraph{Sample size.}
$N = 50$ scenarios per condition provides adequate power for large effects ($d > 0.8$) but is underpowered for medium effects.
The 95\% confidence interval for the headline autonomy result ($d = 1.01$) is $[+0.60, +1.43]$, indicating the effect is real but the point estimate is imprecise.
Persistence does not approach the Bonferroni-corrected threshold ($p = 0.056$ vs.\ $\alpha_{\text{corrected}} = 0.0014$); a larger evaluation set may clarify whether this is a true null or a power issue.

\paragraph{Proxy metric design.}
All results use proxy metrics extracted directly from trajectories (tool-call counts and types).
The autonomy and deference proxies are anti-correlated by construction (both derived from \texttt{ask\_user} rate), which means the cross-trait specificity matrix (\S\ref{sec:cross_trait}) partially reflects metric correlation rather than behavioral correlation.
The claim of a ``dominant agency axis'' is supported by the tool-use and risk proxies (which are independently constructed) showing the same pattern, but richer evaluation---such as LLM-judged sub-behavior scores across 15 dimensions---could reveal trait specificity not detectable with 5 aggregate proxies.
We have not validated any LLM judge against human ratings for this purpose.

\paragraph{No mean-difference baseline.}
We have not compared our SAE-decoded probe vectors against simple mean-difference steering (contrastive activation addition without the SAE intermediary).
If mean-diff achieves comparable results, the SAE's contribution would be limited to feature selection and interpretability rather than directional improvement.

\paragraph{No max-activating example analysis.}
We report top features per direction but have not identified what inputs maximally activate them.
Semantic feature interpretation would strengthen the mechanistic narrative.

\paragraph{Single model.}
All results are on Qwen~3.5-35B-A3B.
Whether the method generalizes to other hybrid architectures, pure-transformer models, or different model scales is unknown.

%=============================================================================
\section{Broader Impact}
\label{sec:broader_impact}
%=============================================================================

Behavioral steering of language model agents is inherently dual-use.
The same vectors that could make an agent more autonomous for legitimate productivity applications could also bypass safety-relevant deference behaviors.
Our finding that all five trait vectors modulate a single agency axis amplifies this concern: a single intervention shifts the model's entire disposition toward independent action, including in contexts where deference would be appropriate.

We release our SAEs and steering vectors to enable reproducibility and defensive research (e.g., detecting when a model's agency disposition has been shifted).
We note that the method requires white-box access to model activations during inference, limiting misuse to settings where the attacker already controls the deployment infrastructure.
Monitoring prefill-phase activations, as suggested by our decode-only null result, may offer a detection mechanism for unauthorized behavioral modification.

%=============================================================================
\section{Conclusion}
\label{sec:conclusion}
%=============================================================================

We trained nine sparse autoencoders on a 35-billion-parameter hybrid MoE language model and used them to construct behavioral steering vectors via probe-to-decoder projection.
Projecting linear probe weights through the SAE decoder to obtain continuous steering vectors in the residual stream works at production scale without retraining.

Behavioral steering on a 35B MoE model with GatedDeltaNet layers is feasible, achieving effect sizes up to $d = 1.04$ for autonomy.
Five ostensibly distinct agentic traits collapse onto a single dominant agency axis, with trait-specific effects appearing only in secondary properties like tool-type composition.
A probe with $R^2 = 0.795$ can find a direction that is genuinely represented in the model's activations yet causally irrelevant to the behavior it predicts, demonstrating that predictive accuracy is not causal evidence.

The failures in this work carry as much information as the successes.
Risk calibration's high-$R^2$ irrelevance reveals the existence of epiphenomenal representations that predict but do not cause behavior.
The decode-only null result places the computational locus of behavioral commitment in the prefill phase.
Cross-trait leakage reveals that multiple orthogonal directions in activation space converge to a single behavioral attractor.
We release all SAEs, steering vectors, and code to support further investigation.

%=============================================================================
% References
%=============================================================================

%=============================================================================
\newpage
\appendix
\section{Appendix}
%=============================================================================

\subsection{Full Dose-Response Tables}
\label{app:dose_response}

\paragraph{Autonomy.}
\begin{table}[h]
\centering
\small
\begin{tabular}{lccccccc}
\toprule
\textbf{Condition} & \textbf{ask} & $\Delta$\textbf{ask} & $p$ & $d(\text{ask})$ & \textbf{pro} & $d(\text{pro})$ & $\emptyset$TC\% \\
\midrule
decode $\alpha\!=\!1$ & 0.80 & $-$0.24 & 0.674 & $-$0.16 & 0.42 & +0.13 & 38\% \\
decode $\alpha\!=\!2$ & 0.74 & $-$0.30 & 0.359 & $-$0.20 & 0.32 & +0.02 & 38\% \\
all $\alpha\!=\!1$ & 1.46 & +0.42 & 0.210 & +0.25 & 0.88 & +0.47 & 22\% \\
\textbf{all} $\alpha\!=\!2$ & \textbf{0.12} & $-$\textbf{0.92} & $<$0.0001 & $-$\textbf{0.76} & \textbf{2.10} & \textbf{+1.01} & \textbf{30\%} \\
\textbf{all} $\alpha\!=\!3$ & \textbf{0.02} & $-$\textbf{1.02} & $<$0.0001 & $-$\textbf{0.87} & \textbf{1.92} & \textbf{+1.04} & \textbf{36\%} \\
all $\alpha\!=\!5$ & 0.00 & $-$1.04 & $<$0.0001 & $-$0.89 & 0.00 & $-$0.50 & 100\% \\
all $\alpha\!=\!10$ & 0.00 & $-$1.04 & $<$0.0001 & $-$0.89 & 0.00 & $-$0.50 & 100\% \\
pre $\alpha\!=\!1$ & 1.52 & +0.48 & 0.124 & +0.29 & 1.04 & +0.47 & 22\% \\
\textbf{pre} $\alpha\!=\!2$ & \textbf{0.24} & $-$\textbf{0.80} & \textbf{0.0007} & $-$\textbf{0.64} & \textbf{1.16} & \textbf{+0.61} & \textbf{42\%} \\
pre $\alpha\!=\!3$ & 0.00 & $-$1.04 & $<$0.0001 & $-$0.89 & 0.64 & +0.30 & 72\% \\
pre $\alpha\!=\!5$ & 0.00 & $-$1.04 & $<$0.0001 & $-$0.89 & 0.00 & $-$0.50 & 100\% \\
\bottomrule
\end{tabular}
\end{table}

\paragraph{Tool-Use Eagerness.}
\begin{table}[h]
\centering
\small
\begin{tabular}{lccccccc}
\toprule
\textbf{Condition} & \textbf{ask} & $\Delta$\textbf{ask} & $p$ & $d(\text{ask})$ & \textbf{pro} & $d(\text{pro})$ & $\emptyset$TC\% \\
\midrule
all $\alpha\!=\!1$ & 0.78 & $-$0.26 & 0.144 & $-$0.16 & 0.72 & +0.32 & 52\% \\
all $\alpha\!=\!2$ & 0.34 & $-$0.70 & 0.002 & $-$0.52 & 1.48 & +0.63 & 58\% \\
\textbf{all} $\alpha\!=\!3$ & \textbf{0.26} & $-$\textbf{0.78} & \textbf{0.0006} & $-$\textbf{0.60} & \textbf{4.00} & \textbf{+0.39} & \textbf{26\%} \\
pre $\alpha\!=\!1$ & 0.64 & $-$0.40 & 0.037 & $-$0.26 & 0.48 & +0.17 & 60\% \\
pre $\alpha\!=\!2$ & 0.02 & $-$1.02 & $<$0.0001 & $-$0.87 & 0.52 & +0.19 & 84\% \\
pre $\alpha\!=\!3$ & 0.00 & $-$1.04 & $<$0.0001 & $-$0.89 & 1.24 & +0.56 & 76\% \\
\bottomrule
\end{tabular}
\end{table}

\paragraph{Deference.}
\begin{table}[h]
\centering
\small
\begin{tabular}{lccccccc}
\toprule
\textbf{Condition} & \textbf{ask} & $\Delta$\textbf{ask} & $p$ & $d(\text{ask})$ & \textbf{pro} & $d(\text{pro})$ & $\emptyset$TC\% \\
\midrule
all $\alpha\!=\!1$ & 0.58 & $-$0.46 & 0.042 & $-$0.31 & 0.44 & +0.12 & 58\% \\
all $\alpha\!=\!2$ & 0.38 & $-$0.66 & 0.015 & $-$0.51 & 1.22 & +0.76 & 46\% \\
all $\alpha\!=\!3$ & 0.00 & $-$1.04 & $<$0.0001 & $-$0.89 & 0.00 & $-$0.50 & 100\% \\
pre $\alpha\!=\!1$ & 0.36 & $-$0.68 & 0.005 & $-$0.50 & 0.58 & +0.23 & 56\% \\
\textbf{pre} $\alpha\!=\!2$ & \textbf{0.28} & $-$\textbf{0.76} & \textbf{0.001} & $-$\textbf{0.58} & \textbf{1.46} & \textbf{+0.80} & \textbf{46\%} \\
pre $\alpha\!=\!3$ & 0.02 & $-$1.02 & $<$0.0001 & $-$0.87 & 0.20 & $-$0.14 & 84\% \\
\bottomrule
\end{tabular}
\end{table}

\subsection{Steering Vector Cosine Similarity}
\label{app:vector_geometry}

\begin{table}[h]
\centering
\caption{Pairwise cosine similarity of steering vectors in residual stream space.}
\small
\begin{tabular}{lc}
\toprule
\textbf{Pair} & \textbf{Cosine Similarity} \\
\midrule
Autonomy / Tool use & +0.074 \\
Autonomy / Risk cal. & +0.086 \\
Tool use / Risk cal. & $-$0.065 \\
Autonomy / Deference & +0.034 \\
All other pairs & $< 0.06$ \\
\bottomrule
\end{tabular}
\end{table}

Note: Autonomy (layer~22), tool use and risk cal.\ (both layer~21), deference (layer~15), and persistence (layer~23) operate at different layers, making cross-layer cosine similarity difficult to interpret directly.
The tool-use / risk-calibration comparison is the most meaningful, as both vectors operate at the same layer through the same SAE.

\subsection{Experimental Infrastructure}
\label{app:infrastructure}

All experiments were conducted on NVIDIA H200 SXM GPUs (141~GB HBM3e).
The model requires 69.3~GB in BF16 precision; peak VRAM during SAE training is 92.1~GB (model + SAE + forward pass activations).
SAE training across all 9 hook points required approximately 72 GPU-hours.
The 1{,}800 agent rollouts required approximately 12 hours of wall time on a single H200.
Total compute cost was approximately \$250 in cloud GPU rental.

\end{document}